\documentclass[a4paper,conference]{IEEEtran}
\IEEEoverridecommandlockouts

\usepackage{cite}
\usepackage{amsmath,amssymb,amsfonts}
\usepackage{graphicx}
\usepackage{booktabs}
\usepackage{url}

\usepackage{algorithmic}
\usepackage[dvipsnames]{xcolor}

\usepackage{subcaption}
\usepackage{multirow}
\newcommand{\authorfont}{\fontsize{11pt}{10pt}\selectfont}
\renewcommand\IEEEkeywordsname{Keywords}

\begin{document}

\title{Aircraft Landing Time Prediction with Deep Learning on Trajectory Images\\\vspace{10pt} 
}

\author{\IEEEauthorblockN{\authorfont Liping Huang, Sheng Zhang, Yicheng Zhang, Yi Zhang, Yifang Yin}
\IEEEauthorblockA{Institute for Infocomm Research, Agency for Science, Technology and Research, Singapore\\
}
}

\maketitle

\begin{abstract}
Aircraft landing time (ALT) prediction is crucial for air traffic management, especially for arrival aircraft sequencing on the runway. In this study, a trajectory image-based deep learning method is proposed to predict ALTs for the aircraft entering the research airspace that covers the Terminal Maneuvering Area (TMA). Specifically, the trajectories of all airborne arrival aircraft within the temporal capture window are used to generate an image with the target aircraft trajectory labeled as red and all background aircraft trajectory labeled as blue. The trajectory images contain various information, including the aircraft position, speed, heading, relative distances, and arrival traffic flows. It enables us to use state-of-the-art deep convolution neural networks for ALT modeling. We also use real-time runway usage obtained from the trajectory data and the external information such as aircraft types and weather conditions as additional inputs. Moreover, a convolution neural network (CNN) based module is designed for automatic holding-related featurizing, which takes the trajectory images, the leading aircraft holding status, and their time and speed gap at the research airspace boundary as its inputs. Its output is further fed into the final end-to-end ALT prediction. The proposed ALT prediction approach is applied to Singapore Changi Airport (ICAO Code: WSSS) using one-month Automatic Dependent Surveillance-Broadcast (ADS-B) data from November 1 to November 30, 2022. Experimental results show that by integrating the holding featurization, we can reduce the mean absolute error (MAE) from 82.23 seconds to 43.96 seconds, and achieve an average accuracy of 96.1\%, with 79.4\% of the predictions errors being less than 60 seconds.
\end{abstract}


\section{Introduction}
The aircraft landing time (ALT) is the time when an aircraft touches down on the runway \cite{c01}. The ALT and its prediction are of great significance for many stakeholders and purposes. For example, airlines can inform passengers about the estimated arrival time. Airports and flight operators can use the ALT information to plan the support services for inbound flights, such as parking, fueling and loading \cite{c02}. Furthermore, accurate predictions of ALT are essential for optimizing arrival aircraft schedules \cite{c05}. The arrival manager (AMAN) systems depend on ALT estimation to sequence the runway slot efficiently \cite{c08}, and to provide precise information (e.g., time to gain, time to lose) for air traffic controllers (ATCOs) to conduct speed control \cite{c09}.

Given the importance of ALT and its prediction, various data-driven machine learning-based approaches have been developed for the ALT prediction. The most commonly used data is the trajectory data because aircraft trajectories of airborne flights contain plenty of information for real-time air traffic. The aircraft trajectories are helpful to extract the aircraft location, speed, traffic context, relative spacing between aircraft, etc \cite{c012}. The study in \cite{c013} uses airborne trajectories in the extended terminal maneuver airspace of WSSS to obtain the latitude, longitude, altitude, speed, rate of climb, heading of the aircraft and the entry zone of an aircraft for ALT prediction. The study in \cite{c014} explores another set of features from the trajectories for modeling ALTs, including the calculated arrival pressure, and the calculated sequencing pressure. Meanwhile, researchers in \cite{c015} obtain the traffic densities at different granularities from the trajectory data. Though various information from aircraft trajectories is extracted for the ALT prediction, these studies do not involve aircraft holding stages, which makes the ALT prediction more challenging.
    
The most commonly applied machine learning methods for ALT prediction are the decision tree-based models \cite{c015, c016}. For instance, the study in \cite{c015} tested several decision tree-based machine learning models for the ALT predictions, including random forest and the gradient boosting machine(GBM). The performance of these kind of models depends on the features, which requires amounts of feature engineering work. Data missing may make the models lose efficacy since missing data is removed as in \cite{c017}. Besides, manual feature engineering-based methods\cite{c018} depend on the data quality and human knowledge of the research problem. Conversely, deep learning-based methods can help capture the unseen patterns of the data. The research in \cite{neural} proposes a deep learning-based method for ALT prediction, where the the time series-based deep learning model, LSTM, is used on the non-linear (tanh) combination of the raw trajectory's latitude, longitude, altitude, and speed. Unlike CNNs, the time-series based model cannot directly capture the spatial context among aircraft in TMA, e.g., the distances between aircraft.

\begin{table*}[t]
  \begin{center}
    \caption{Acronym (Variable) and Corresponding Definition}
    \label{tab_acronyms}
    \begin{tabular}{|c|c|c|}
    \hline
      \textbf{Acronym (Variable)} &\textbf{Definition} & \textbf{Description} \\
      \hline
      TRC & TMA Research Airspace Circle & 50 NM to Changi Airport\\
      \hline
      TBX & TMA Extended Boundary & 60 NM to Changi Airport\\
      \hline
      TBE & between TRC and TBX & flying distance is around 10 NM\\
      \hline
      $\mathcal{T}^{j}_{\scalebox{0.7}{TRC}}$ & timestamp when an aircraft $j$ arrives at TRC & calculated from ADS-B data\\
      \hline
      $\mathcal{T}^{j}_{\scalebox{0.7}{THR}}$ & timestamp when an aircraft $j$ arrives at the runway threshold & calculated from ADS-B data\\
      \hline
      $\mathcal{T}^{\mathcal{L}(j)}_{\scalebox{0.7}{TRC}}$ & timestamp of leading aircraft of $j$ arrives at TRC& $\mathcal{L}(j)$ is the leading one of $j$ when $j$ arrive at TRC\\
      \hline
      $\mathbf{t}_j$ & landing time for aircraft $j$ as $\mathcal{T}^{j}_{\scalebox{0.7}{THR}}-\mathcal{T}^{j}_{\scalebox{0.7}{TRC}}$ & target prediction variable\\
      \hline
      
      $v^j_{\scalebox{0.7}{E}}$  & mean speed of aircraft $j$ within TBE & in knots, calculated from ADS-B \\
      \hline
      $v^{\mathcal{L}(j)}_{\scalebox{0.7}{E}}$  & speed of the leading aircraft of $j$ within TBE & in knots, calculated from ADS-B \\
      \hline
      $\overline{v}_{\scalebox{0.7}{E}}$  & the average $v^j_{\scalebox{0.7}{E}}$ values for all aircraft within TBE & in knots, calculated from ADS-B \\
      \hline
      $\tau$ & trajectory image capture window with size $\tau$ seconds & time window as [$\mathcal{T}^{j}_{\scalebox{0.7}{TRC}}-\tau$, $\mathcal{T}^{j}_{\scalebox{0.7}{TRC}}$]\\
      \hline
       $\delta$ & tabular feature capture window with size $\delta$ seconds & time window as [$\mathcal{T}^{j}_{\scalebox{0.7}{TRC}}-\delta$, $\mathcal{T}^{j}_{\scalebox{0.7}{TRC}}$]  \\
      \hline
      $X_j^{\tau_j}$ & generated trajectory image for aircraft $j$ in capture window $\tau$ & used to represent TMA context  \\
      \hline
      $X_j^{\delta_j}$ & tabular features for aircraft $j$ captured in window $\delta$ & runway usage, weather, aircraft type, etc. \\
      \hline
    
    \end{tabular}
  \end{center}
\end{table*}

In this study, we propose using trajectory images instead of raw aircraft latitude, longitude, speed, heading information for ALT modeling. Specifically, we define a TMA research airspace circle (TRC) that is centered around the airport and covers the TMA for the problem formulation. For the case study of Singapore Changi Airport, the TRC has a radius of 50 nautical miles (NM), which is sufficient to cover the TMA of the airport since its TMA is located at a distance of 40NM from the airport. For each target aircraft that arrives at TRC, we generate a trajectory image with the target aircraft trajectory colored in red and the background aircraft trajectory colored in blue. The trajectory images inherently contain the aircraft's location, speed, heading, relative distance, entry zone, and arrival traffic flow in the TRC airspace. Based on the generated trajectory images, we design an end-to-end deep learning model that leverages state-of-the-art convolution neural networks for predicting aircraft landing times. Contributions of this study are summarized as follows:
\begin{itemize}
    \item  Trajectory images are proposed to be generated and utilized in the ALT modeling. These images can capture the aircraft's position, speed, heading, entry zone, arrival traffic flow, and relative distances among airborne flights. The trajectory image generation can reduce our major efforts in the complicated feature engineering process and enable us to apply state-of-the-art convolution neural networks for the ALT modeling problem. 
    \item A CNN-based module is designed for the automatic holding-related featurization. Except for trajectory images, this module also takes the holding status of the leading aircraft, the time/speed gap at TRC with the leading aircraft, and the speed variation at TRC as its inputs. Its output is further fed into the final ALT prediction to make it capable of capturing the holding-related features, and hence an end-to-end ALT prediction model is proposed.
    
    \item The proposed ALT prediction approach has been testified with the case study of Singapore Changi Airport, where one-month ADS-B data from November 1 to November 30, 2022 has been used. Experimental results show that incorporating the holding featurization module reduced the mean absolute error (MAE) from 82.23 seconds to 43.96 seconds. The bad prediction ratio dropped by 1.31 percent. Furthermore, 79.4\% of the prediction errors are less than 60 seconds. 
\end{itemize}

The rest of this paper is organized as follows. Section \ref{pre} presents the methodology. Section \ref{exp} presents the experimental evaluations and corresponding analysis of the prediction accuracy. Conclusions and discussion for future work are finally drawn in Section \ref{conclusion}.

\section{Methodology}
\label{pre}

\subsection{Problem Statement}
For clear and concise expression, acronyms, variables, and corresponding definition used in this study are summarized in Table \ref{tab_acronyms}. In this study, we focus on predicting the landing time of each inbound aircraft when it arrives at TRC of Singapore Changi Airport. The landing time $\mathbf{t}_j$ for an aircraft $j$ is the time calculated as $\mathbf{t}_j=\mathcal{T}^{j}_{\scalebox{0.7}{THR}}-\mathcal{T}^{j}_{\scalebox{0.7}{TRC}}$, where $\mathcal{T}^{j}_{\scalebox{0.7}{THR}}$ and $\mathcal{T}^{j}_{\scalebox{0.7}{TRC}}$ are respectively the timestamps when the aircraft $j$ arrives at the runway threshold and the TRC (50NM to the airport). Both timestamps are extracted from aircraft trajectory data. For predicting $\textbf{t}_j$, we intend to construct a mapping function $\textbf{t}_j=\mathcal{F}(X_j^{\tau, \delta})$. In this study, we utilize deep neural networks to construct the mapping function, and the input feature $X_j^{\tau, \delta}$ is a combination of the generated trajectory image $ X_{j}^{\tau_{j}}$ within the capture window [$\mathcal{T}^{j}_{\scalebox{0.7}{TRC}}-\tau$, $\mathcal{T}^{j}_{\scalebox{0.7}{TRC}}$], and the tabular features $X_{j}^{\delta_{j}}$ captured in the time window [$\mathcal{T}^{j}_{\scalebox{0.7}{TRC}}-\delta$, $\mathcal{T}^{j}_{\scalebox{0.7}{TRC}}$]. Hence, the objective is to construct the mapping function 

\begin{equation}
    \mathbf{t}_{j} = \mathcal{F}(X_{j}^{\tau_{j}}, X_{j}^{\delta_{j}}), j\in\mathcal{J}
\end{equation}

where $X_{j}^{\tau_{j}}, X_{j}^{\delta_{j}}$ are respectively the trajectory image and tabular features for aircraft $j$. In the following subsections, we elaborate on the methodology overview, dataset description, data analysis, feature engineering (including the trajectory image generation and the related tabular feature extraction), holding featurization, and deep convolution neural networks for the proposed approach.

\begin{figure*}[!ht]
\begin{center}
	\includegraphics[width=0.78\textwidth]{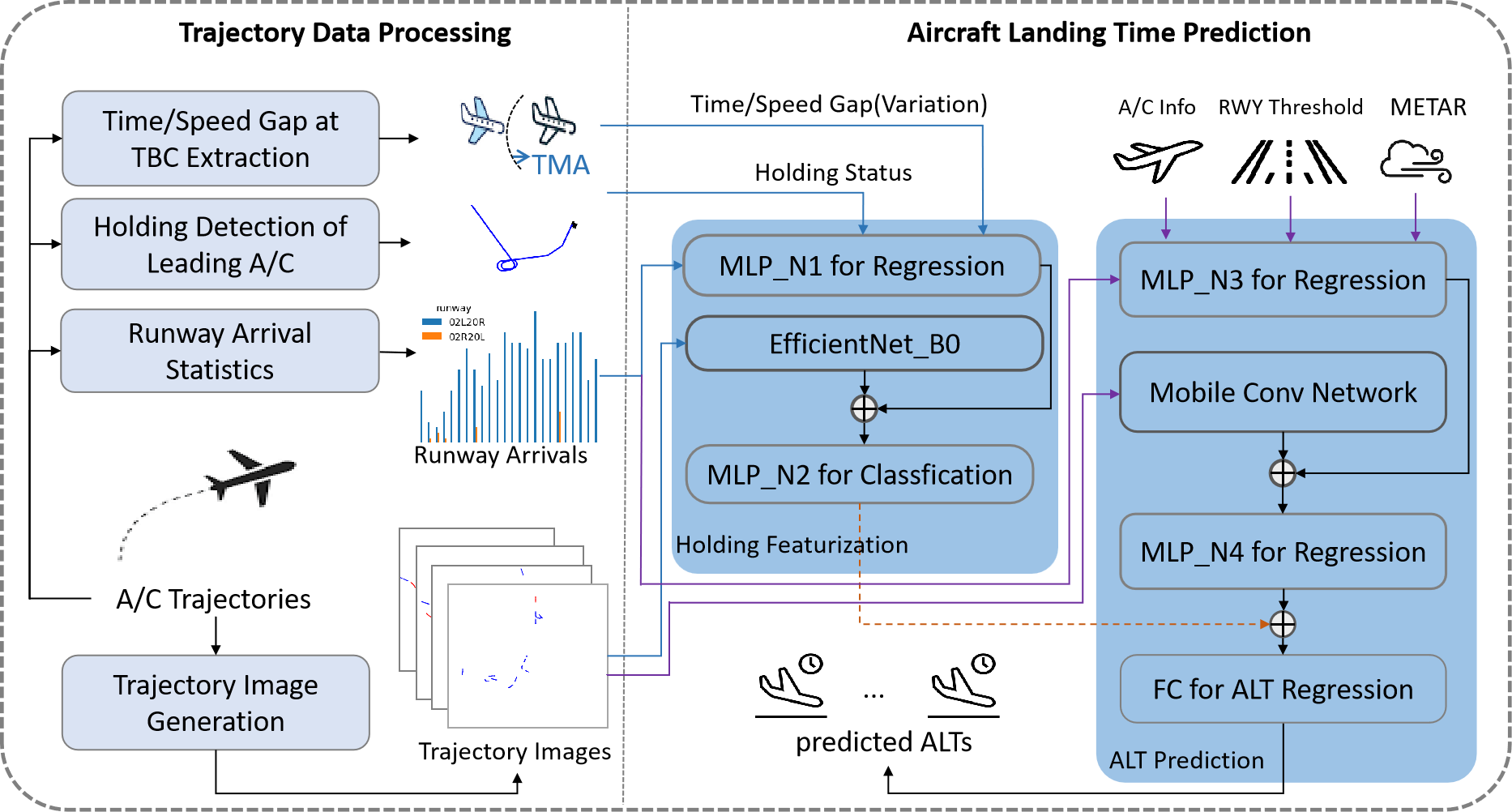}
	\caption{Overview of the Proposed Method.}
	\label{framework}
 \end{center}
\end{figure*}

\subsection{Methodology Overview}
The framework for the proposed aircraft landing time prediction method is described in Fig. \ref{framework}. The ALT prediction takes the generated trajectory image, the aircraft information, runway threshold usage, weather condition, and the output from the Holding Featurization as its inputs. Historical trajectories of aircraft are used to generate the trajectory images and extract the runway arrivals, which are the inputs for both the Holding Featurization and the ALT Prediction. Additionally, the leading aircraft holding status, the time gap and speed gap with the leading aircraft at TRC, and the speed variation at TRC are used as the inputs of the Holding Featurization.

\subsection{Dataset}
\subsubsection{Trajectory Data}
\begin{figure}[!hb]
\begin{center}
	\includegraphics[width=0.8\columnwidth]{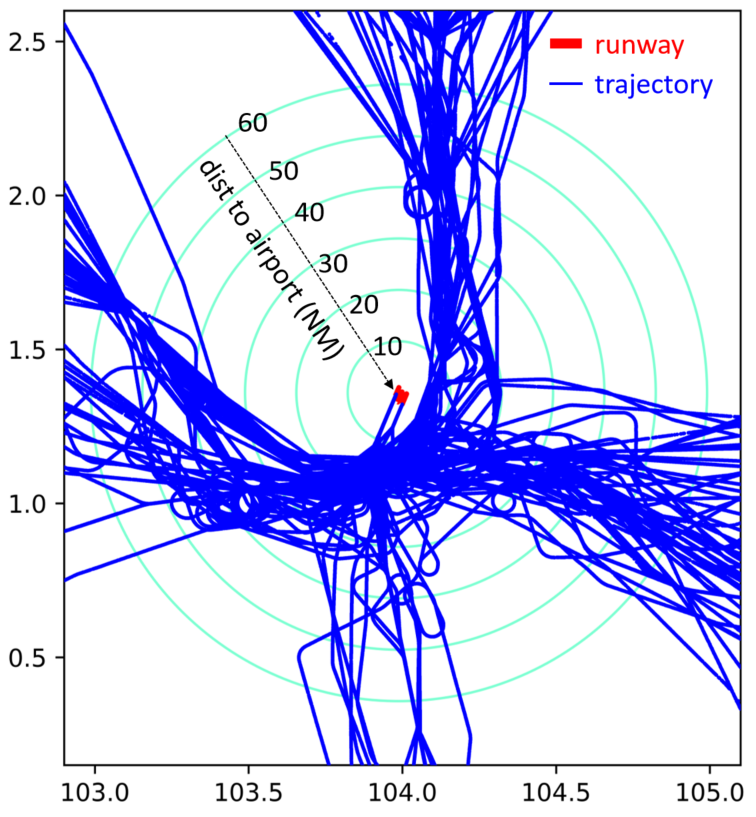}
	\caption{Arrival Aicraft Trajectories in One Day for Changi Airport.}
	\label{trajectory}
 \end{center}
\end{figure}

The trajectory data source of this study is Automatic Dependent Surveillance-Broadcast (ADS-B) flight data. One-month ADS-B data from November 1 to November 30, 2022,  have been used in this paper. Even though the ADS-B dataset is of high quality, where the data sampling granularity is typically one positioning point per second, some of the data segments have slight missing points, hence a quadratic interpolation is used to impute the missing points for both latitude and longitude. By matching the imputed ADS-B positioning points to Changi airport runways, we extract the arrival aircraft data and get the corresponding landing runway for each aircraft. After removing the outliers (outside 3 times stand deviation to the mean value), we have 8396 aircraft for this study. Since the TRC is 50NM from the airport, the longitude range is (103.0, 105.0) and the latitude range is (0.5, 2.25). Fig. \ref{trajectory} shows the one day's trajectories of arrival flights with the distance circles. 

\subsubsection{Meteorological Data}
the meteorological data (METAR) for Singapore Changi airport station are obtained from Iowa Environmental Mesonet from Iowa State University \cite{c021}, where the wind direction, wind speed in knots, visibility in miles, wind gust in knots, sky coverage and the altitude in feet are provided.
   
\subsubsection{Flight Plan and Online Aircraft Performance Database}
    Flight Plan (FPL) and online aircraft performance database (APD) \cite{c022} are used to obtain the aircraft RECAT-EU. Each record in the FPL contains the aircraft model type (e.g., A320), and the online APD provides the mapping from model type to the RECAT-EU. The aircraft are classified into six categories of RECAT-EU, namely, Light, Lower Medium, Upper Medium, Lower Heavy, Upper Heavy, and Supper Heavy. Our dataset contains five of them, as "light" aircraft did not land at Singapore Changi Airport in November 2022.


\subsection{Data Analysis}


\subsubsection{Landing Time Analysis}

\begin{figure}[t]
\begin{center}
	\includegraphics[width=0.78\columnwidth]{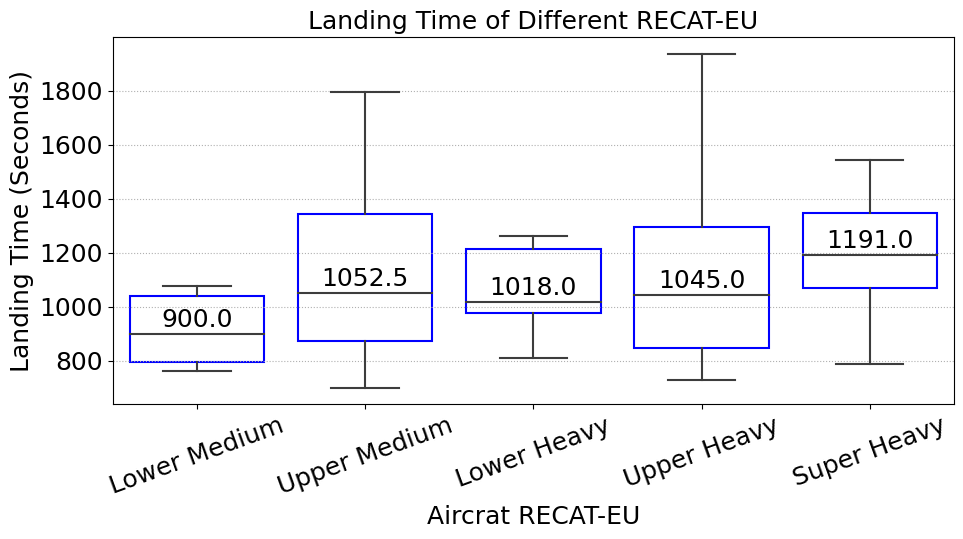}
	\caption{Aircraft Landing Time vs RECAT-EU.}
	\label{eta_recat}
 \end{center}
\end{figure}

As described in the problem statement section, the landing time $\mathbf{t}_j$ for aircraft $j$ is calculated as the time gap between the timestamp for landing at the runway threhold $\mathcal{T}^{j}_{\scalebox{0.7}{THR}}$, and the timestamp $\mathcal{T}^{j}_{\scalebox{0.7}{TRC}}$ when the aircraft arrives at TRC, where both timestamps are extracted from the ADS-B trajectory data. The landing time distribution for all aircraft categories during a day is shown in Fig. \ref{eta_recat}. It indicates that "Super Heavy" aircraft tends to have longer landing times, and "Lower Medium" aircraft tends to have the shorter landing times. Hence, it's essential to involve the RECAT-EU of an aircraft in the ALT prediction modeling.

\subsubsection{Runway Threshold Usage for Aircraft Landing}
\begin{figure}[ht]
\begin{center}
	\includegraphics[width=0.85\columnwidth]{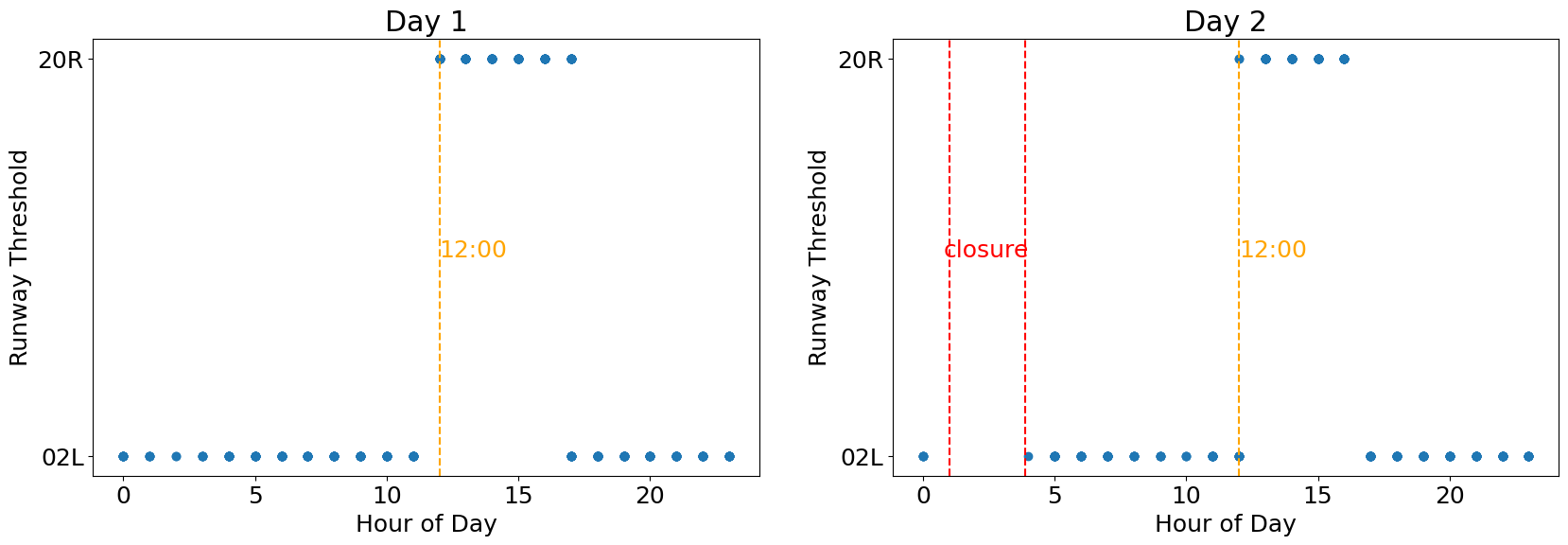}
	\caption{Runway Threshold Usage in Two Days.}
	\label{thr}
 \end{center}
\end{figure}

Specific to the runway 02L20R, the runway threshold used during a day can be changed due to the runway operation (e.g., due to wind direction change, etc.). The runway thresholds for the physical runway 02L20R are 02L and 20R. The actual runway threshold usage for aircraft landing in two days are shown in Fig. \ref{thr}. The figure shows that the runway threshold 02L is used before 12:00 on Day 1 and changed to 20R at 12:00. On Day 2, the runway is closed during the early morning and changed from 02L to 20R between 12:00-13:00. The runway threshold changes for landing can cause extra waiting time for aircraft in the TMA. Accordingly, we use a runway change label (0/1) in the ALT modeling.

\subsubsection{Holding Detection and Impact on ALT}
\begin{figure}[!ht]
\begin{center}
	\includegraphics[width=0.88\columnwidth]{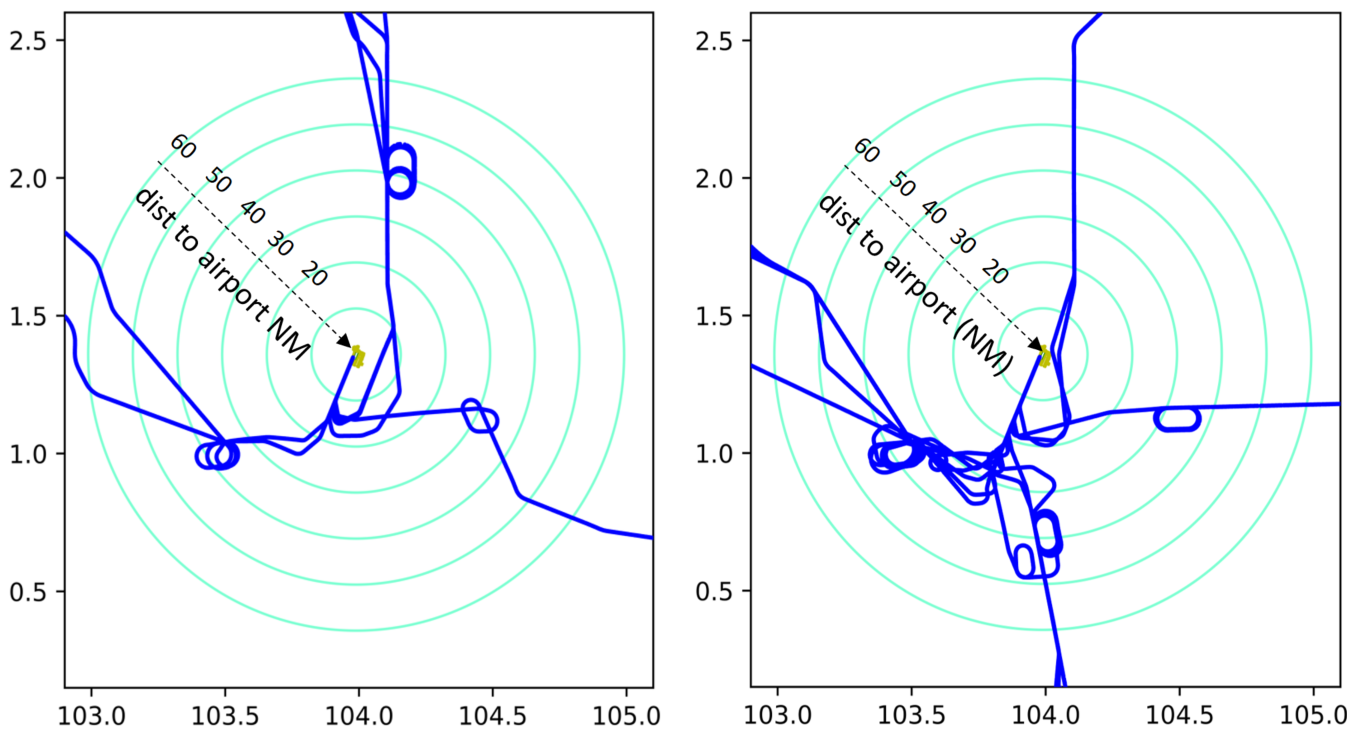}
	\caption{Trajectories with Holdings in Two Days.}
	\label{holding_traj}
 \end{center}
\end{figure}

An aircraft may have to enter a holding stage due to the air traffic control, which makes the landing time prediction more difficult. In this study, we propose to integrate the holding potentials of each aircraft for the aircraft landing time prediction. In this section, we will analyze the aircraft holdings and their effect on the ALTs.
The holding detection is based on the ADS-B data \cite{d_hld}. The trajectories with holdings in two days are shown in Fig. \ref{holding_traj}. We can note that arrival flights can enter the TMA of Changi airport from four directions, and in each direction, flights may enter a holding stack to follow the air traffic control. The holdings mostly happen between 30NM and 50NM to the airport, and we set the TRC as 50NM to cover these holdings. 

\begin{figure}[htb]
\begin{center}
	\includegraphics[width=0.65\columnwidth]{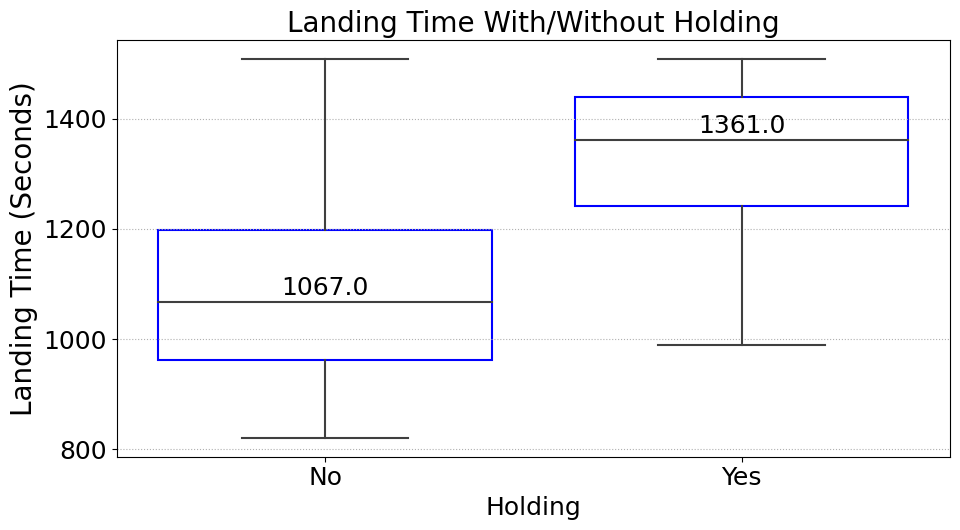}
	\caption{Aircraft Landing Times vs Holdings.}
	\label{eta_hld}
 \end{center}
\end{figure}

To illustrate the effect of holdings on the ALTs, we compare the ALTs for aircraft with and without holdings in Fig. \ref{eta_hld}. It shows that flights with holdings have higher ALTs than the flights without holdings. For this reason, we propose to utilize a module to capture the holding probability-related features for the ALT prediction. The following subsection presents the input features for the proposed approach.

\begin{table*}[tb]
    \caption{Features for ALTs and Their Representations.}
    \centering
    \begin{tabular}{|c|c|c|c|c|}
    \hline
        \textbf{Indicator} &  \textbf{Description} & \textbf{Representation}  & \textbf{Feature Category} & \textbf{Data Source} \\
        \hline
        aircraft position  &  pixel positions in the image  &                  &       &     \\
         \cline{1-2}
        aircraft speed  &  trajectory length in the image  &           &      &      \\
         \cline{1-2}
        aircraft heading & tail to head direction  & trajectory image  & TMA Context & ADS-B\\
         \cline{1-2}
        entry zone & pixel positions of the target aircraft & & &\\
         \cline{1-2}
        arrival traffic & number of trajectories in the image & & &\\
         \cline{1-4}
        dynamic runway setting & runway threshold change label & 0/1 & Runway Operation & \\
        \cline{1-3}
        runway usage & arrivals on each runway & real number &   &  \\
        \hline
        Weather & drct, sknt, gust, vsby, skyl1, skyc1  & real number (in knots), 0/1 &  & METAR\\
        \cline{1-3}
        \cline{5-5}
        seasonality & is\_peakhour, is\_weekday & 0/1 & External Info &Flight Plan/APD \\
        \cline{1-3}
        A/C info & Aircraft RECAT-EU & 0,1,2,3,4,5 &  & \\
    \hline    
    \end{tabular}    
    \label{table_feature}
\end{table*}



\begin{figure*}[htb]
    \centering
    \begin{subfigure}[b]{0.42\textwidth}
      \centering
    	\includegraphics[width=0.42\textwidth]{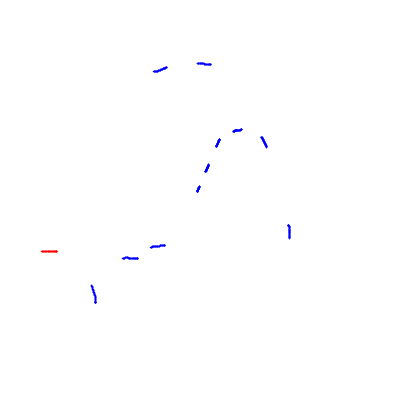}
           \caption{Trajectory Image Example, Landing from West}
       \label{tra_a}
   \end{subfigure}
   \begin{subfigure}[b]{0.42\textwidth}
      \centering
    	\includegraphics[width=0.42\textwidth]{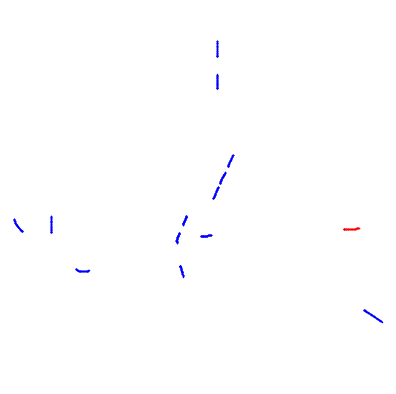}
           \caption{Trajectory Image Example, Landing from East}
       \label{tra_b}
   \end{subfigure}
    \begin{subfigure}[b]{0.42\textwidth}
      \centering
    	\includegraphics[width=0.42\textwidth]{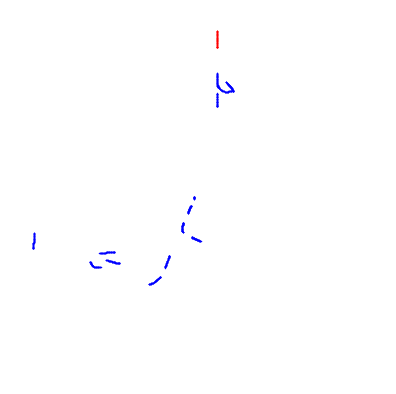}
           \caption{Trajectory Image Example, Landing from North}
       \label{tra_c}
   \end{subfigure}
    \begin{subfigure}[b]{0.42\textwidth}
      \centering
    	\includegraphics[width=0.42\textwidth]{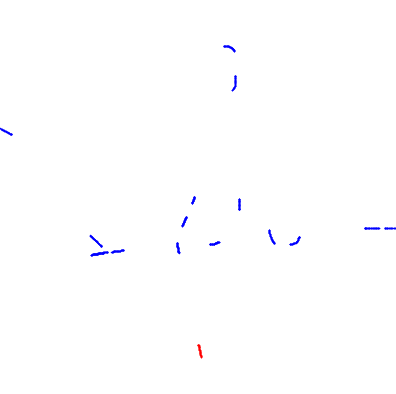}
           \caption{Trajectory Image Example, Landing from South}
       \label{tra_d}
   \end{subfigure}
\caption{Examples of Generated Trajectory Images for Aircraft Landing in Four Different Directions, $\tau=60, \delta=10$.}
\label{traj_img}
\end{figure*}

\subsection{Feature Engineering}
\subsubsection{Factors and Representation}
As we analyzed in the previous subsection, various factors can affect $\mathbf{t}_{j}$, such as the TMA traffic context, the weather conditions, the runway settings and operations, as well as the control intent for safety guarantees. In this study, we explore how to integrate these factors for predicting the aircraft landing time $\mathbf{t}_{j}$. We summarize the potential infecting factors in Table \ref{table_feature}. Three categories of features are used: the TMA context, the runway operation, and the external information. The TMA context is represented as the generated trajectory image, which is generated from the ADS-B data to capture the aircraft position, speed, heading, entry zone, and the TMA arrival traffic. We will describe each type of features in the follow subsections.

\subsubsection{Trajectory Image Generation}
In this study, we generate the trajectory image for representing the TMA context. Given the capture window size $\tau$, we generate the trajectory image for an aircraft $j$ when it arrives at TRC. The track position points for all aircraft inside TRC and within the time window [$\mathcal{T}^{j}_{\scalebox{0.7}{TRC}}-\tau$, $\mathcal{T}^{j}_{\scalebox{0.7}{TRC}}$] are collected to plot the image. Specifically, the trajectory of the target aircraft $j$ are marked in red, and all other planes' trajectories are marked in blue.   

Fig. \ref{traj_img} shows examples of the generated images for aircraft landing from four different directions to Singapore Changi Airport. The image generation is based on the ADS-B data, which provides one point per second for each aircraft (missing data points are imputed with interpolation). For a given $\tau$ value, all aircraft in one image have the same flying time, so the relative length of the trajectory indicates the aircraft speed. That is, longer trajectory in one image means a higher speed of the aircraft. Moreover, the aircraft heading is determined by the direction from the last point to the first point of the trajectory, where the first point is usually closer to the runway. By generating such a trajectory image, we can capture several kinds of TMA context simultaneously in the image as described in Table. \ref{table_feature}: the aircraft position, speed, heading, the entry zone, relative distances. Additionally, the number of trajectories in the image directly represents the arrival traffic flow in the capture window.

\subsubsection{Runway Operation Feature}
We use the ADS-B data to determine the landing runway and runway threshold for each aircraft. We also count the number of arrival aircraft for each runway within the tabular feature capture window [$\mathcal{T}^{j}_{\scalebox{0.7}{TRC}}-\delta$, $\mathcal{T}^{j}_{\scalebox{0.7}{TRC}}$]. Furthermore,
we set the runway threshold change label to 1 if the runway threshold is changed during the capture window [$\mathcal{T}^{j}_{\scalebox{0.7}{TRC}}-\delta$, $\mathcal{T}^{j}_{\scalebox{0.7}{TRC}}$], and to 0 otherwise.

\subsubsection{External Info}
We obtain the aircraft model type, e.g., A320, from the flight plan dataset and use it to get the corresponding model RECAT-EU type, e.g., Upper Medium, from the online APD. Then we encode the RECATEU type as a number from 0 to 5 (0 for light, 5 for super heavy), and use it as one dimension of the input feature for $X_j^{\delta_j}$. For the weather-related features, we utilize the wind direction (drct), wind speed (sknt), visibility (vsby) in miles, wind gust in knots, sky level 1 altitude in feet, and the sky level 1 coverage, which are all from the METAR dataset. We extract all these features, as well as the seasonality features, within the capture window [$\mathcal{T}^{j}_{\scalebox{0.7}{TRC}}-\delta$, $\mathcal{T}^{j}_{\scalebox{0.7}{TRC}}$], and each is set as one dimension of $X_j^{\delta_j}$ for the ALT modeling.

\subsection{Holding Featurization}
As shown in Fig. \ref{framework}, we design a module to estimate the holding potential of an aircraft for the ALT prediction. The Holding Featurization module takes trajectory images and other tabular features as its input. In this section, we describe the tabular features for the Holding Featurization module. These include the runway arrivals that are also the input for the ALT Prediction module. Besides, the time gap and speed of an aircraft with its leading aircraft at TRC,  and the variation of the speed compared to the average speed within TBE are used. As Fig. \ref{tra_c} illustrates, the target aircraft's leading aircraft enters the holding status, which have a high probability of making the target aircraft also enter the holding stack. We explain these features in detail below. 
\begin{itemize}
    \item the arrivals on runways: the number of aircraft entering holding stage increases when the the runways are busy. During this period, parallel landing can be conducted.
    \item The time gap $\Delta{\mathcal{T}^{j, \mathcal{L}(j)}} = \mathcal{T}^{j}_{\scalebox{0.7}{TRC}} - \mathcal{T}^{\mathcal{L}(j)}_{\scalebox{0.7}{TRC}}$, denoting the the time gap between the target and its leading aircraft arriving at the TRC, here $\mathcal{L}(j)$ is the leading aircraft of aircraft $j$ when aircraft $j$ arrives at TRC;
    \item the speed variation $\Delta v^{j, avg}_{\scalebox{0.7}{E}} = v^j_{\scalebox{0.7}{E}}-\overline{v}_{\scalebox{0.7}{E}}$, which is the gap between the speed of aircraft $j$ with the average speed of all aircraft in TBE, where $\overline{v}_{\scalebox{0.7}{E}}$ is the average speed of all aircraft in TBE.
    \item The speed gap $\Delta v^{j, \mathcal{L}(j)}_{\scalebox{0.7}{E}} = v^j_{\scalebox{0.7}{E}}-v^{\mathcal{L}(j)}_{\scalebox{0.7}{E}}$, which representing the speed difference between aircraft $j$ and its leading aircraft $\mathcal{L}(j)$ within the 10 NM in TBE. 
    \item the holding status of the leading aircraft $\mathcal{L}(j)$, which is a 0/1 label. 
\end{itemize}

\begin{figure}[t]
   \begin{center}
	\includegraphics[width=\columnwidth]{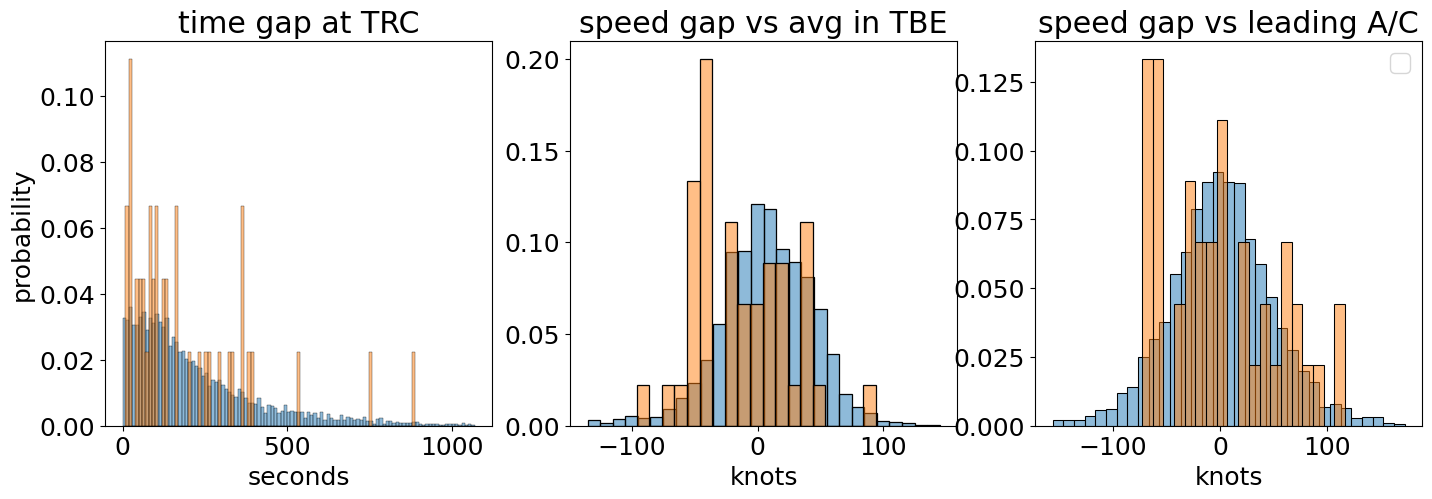}
	\caption{Statistical probability of time gap at TRC, speed gap vs average (avg) speed in TBE, and the speed gap vs its leading aircraft's speed in TBE. \textcolor{ProcessBlue}{ Blue: without holding}, \textcolor{BurntOrange}{Orange: with holding }.}
	\label{hF_sta}
 \end{center}
\end{figure} 

The statistical probability of $\Delta{\mathcal{T}^{j, \mathcal{L}(j)}}$, $\Delta v^{j, avg}_{\scalebox{0.7}{E}}$, and $\Delta v^{j, \mathcal{L}(j)}_{\scalebox{0.7}{E}}$ are shown in Fig. \ref{hF_sta}, which is subject to the with holding and without holding aircraft. It indicates that holding aircraft tend to have smaller values in all three features. Hence, these features are extracted and set as the input for the holding featurization module. Besides, the holding status of a target aircraft's leading aircraft (0/1) is set as an additional input.

\subsection{Deep Convolution Networks}
\subsubsection{Mobile Convolution Network for ALT Prediction}

\begin{figure}[b]
   \begin{center}
	\includegraphics[width=0.5\columnwidth]{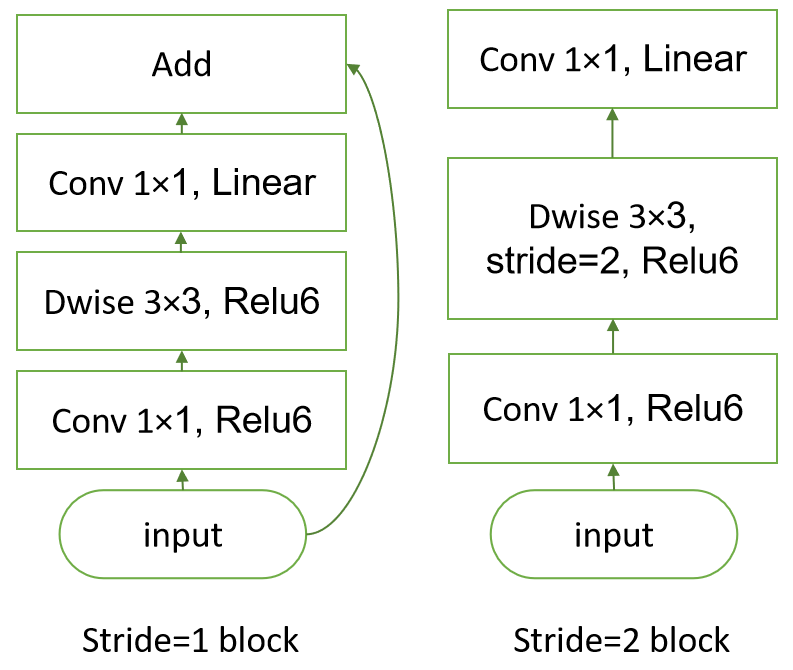}
	\caption{Two kinds of blocks in MobileNetV2 \cite{c025}.}
	\label{mobilenet}
 \end{center}
\end{figure} 

The MobileNetV2 has better efficiency since it utilizes depthwise separable convolutions, which are the key building blocks for many efficient neural network architectures \cite{c025}. Assume that the dimension of the input tensor $\emph{L}_i$ in layer $i$ is $h_i \times w_i \times d_i$, the dimension for the output tensor $\emph{L}_j$ is $h_i \times w_i \times d_j$, and then the standard convolution's kernel size is $\emph{K}\in \mathbb{R}^{k \times k \times d_i \times d_j}$, which require the computation cost of $h_i \cdot w_i \cdot d_i \cdot d_j \cdot k \cdot k$. Whereas, for depthwise separable convolutions, the computation cost is only $h_i \cdot w_i \cdot d_i (k^2 + d_j)$, which reduces computation compared to standard convolution by almost a factor of $k^2$. Generally, the kernel size is $3$, then it means for the computation cost of depthwise separable convolution is 8 to 9 times smaller than that of standard convolutions \cite{c025}.

The main innovation of MobileNetV2 is that it is an inversed residual network \cite{c025}, which means the residual connection is used only when the stride is 1 and the number of channels remains the same. Two kinds of blocks in MobineNetV2 are shown in Fig. \ref{mobilenet}. Depth-wise convolution uses the kernel $3\times 3$, and point-wise convolution uses $1\time 1$ convolution. Between the residual connections, it contains the decompression phase, where the output channels are expanded among connected layers. In this study, we use the default settings for the expanding ratio, i.e., 6, as in the original code \cite{c025}.

\subsubsection{Efficientnet for Holding Probability}
\begin{figure}[t]
   \begin{center}
	\includegraphics[width=\columnwidth]{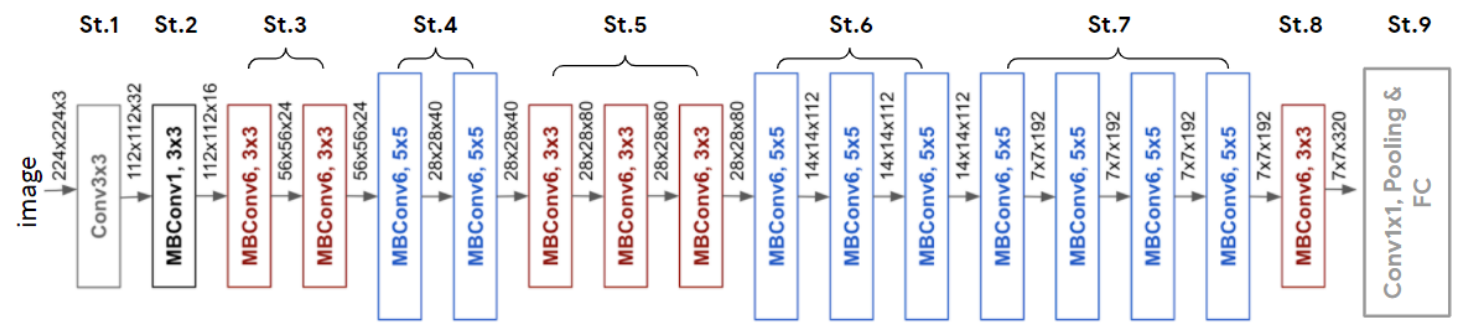}
	\caption{Architecture of EfficientNet-B0 \cite{c026}.}
	\label{effnet}
 \end{center}
\end{figure}

The holding probability is as part of inputs for the landing time prediction, which is captured by the Efficientnet \cite{c026} with the trajectory image as its input. The Efficientnet proposes a new compound scaling method, which uses a compound coefficient $\phi$ to uniformly scales convolution network's width, depth, and resolution in the following principled way \cite{c026}:

\begin{equation} \label{eql}
\begin{split}
    depth: d = \alpha^{\phi} \\
    width: w = \beta^{\phi} \\
    resolution: r= \gamma^{\phi}\\
    s.t., \alpha\cdot\beta^2\cdot\gamma^2 \approx 2\\
    \alpha \geq 1, \beta\geq 1, \gamma \geq 1
\end{split}
\end{equation}

where $\alpha, \beta, \gamma$ are constants that can be determined by a small grid search. $\alpha\cdot\beta^2\cdot\gamma^2 \approx 2$ is to make the  FLOPS (floating point operations per second) approximately increase by $2^\phi$. In this study, we utilize the EfficientNet-B0, where the parameters are $\phi=1, \alpha = 1.2, \beta=1.1, \gamma=1.15$, and the main building block is the mobile inverted bottleneck MBConv. Its architecture is shown in Fig. \ref{effnet}.

\section{Experimental Results and Analysis}
\label{exp}

\begin{table*}[!ht]
\caption{Prediction Accuracy Evaluation of Each Model for All Scenarios.}
    \label{etapred}
    \centering
    \begin{tabular}{c|c|c|c|c|c|c|c|c|c}
 \hline

  \multirow{3}{*}{Metrics}   &  \multirow{3}{*}{$\delta$(in Mins)}  & \multicolumn{2}{c}{MAPE}  & \multicolumn{2}{|c}{MAE}  & \multicolumn{2}{|c}{RMSE}  & \multicolumn{2}{|c}{BadRatio ($\gamma=0.3$)}\\
  \cline{3-10}
   & & $\tau=60$ & $\tau=90$ & $\tau=60$ & $\tau=90$ & $\tau=60$ & $\tau=90$ & $\tau=60$ & $\tau=90$  \\
\hline

                    &10 & \textbf{0.0738} & 0.0899  &  \textbf{82.23}  &  102.04  &  \textbf{115.91} & 140.44  &  \textbf{0.0156} & 0.0175\\
     Baseline        &15 & 0.0789  & 0.0829  &  86.10  &  91.38   &  116.17 & 124.81  &  0.0166 & 0.0191\\
     (No holding Module)  &20 & 0.0808  & 0.0799  &  88.11  &  89.28   &  120.87 & 123.76  &  0.0216 & 0.0175 \\
                    &25 & 0.0767  & 0.0786  &  86.81  &  87.05   &  123.28 & 119.23  &  0.0183 & 0.0158 \\
                    &30 & 0.0788  & 0.0974  &  87.22  &  101.23   &  120.65 & 130.20  &  0.0166 & 0.0308 \\
\hline \hline

                   &10 & \textbf{0.0390} & 0.0759 & \textbf{43.96}   &  86.76    &  \textbf{71.91}   &  123.95  & \textbf{0.0025}  &  0.0208  \\
                   &15 & 0.0762 & 0.0766  & 84.86  &  85.58    &  116.12  &  116.89  & 0.0066  &  0.0150  \\
  Proposed Method  &20 & 0.0859 & 0.0740  & 94.36  &  84.89    &  126.61  &  121.13  & 0.0250  &  0.0125 \\
                   &25 & 0.0763 & 0.0727  & 86.23  &  81.95    &  119.45  &  115.40  & 0.0108  &  0.0141 \\
                   &30 & 0.0772 & 0.0786  & 86.05  &  88.64    &  118.97  &  123.98  & 0.0141  &  0.0175 \\         
\hline
\end{tabular}
\end{table*}

\subsection{Experimental Settings}

We randomly split the total 8396 aircraft in the dataset $\mathcal{D}$ into 70\%, 15\%, 15\% for model training, validation, and testing, respectively, where features are normalized first. We set the loss function as the L1 loss, i.e., the Mean Average Error (MAE), the optimizer as Adam, the batch size as 64, the epoch as 1000,  and the learning rate as 0.001. The network has 6.4M parameters in total.

For the parameters of EfficientNet\_B0, all parameters are set as in \cite{c026}. 
For MobileNetV2, a pretrained mode is used, and all parameters are as in \cite{c025}. Since MobileNetV2 \cite{c025} is designed for classification problems, however, our problem is a regression problem, hence, we use these networks for feature engineering and the final classification layer is replaced by following sequential layers: Linear(1280, 64), BatchNorm1d(64) LeakyReLU(Linear(128)), Dropout(0.1). 
The MLPs in the proposed method as shown in Fig. \ref{framework} are structured as following sequential layers.
(1) MLP\_N1: Linear(5, 16), LeakyReLU.
(2) MLP\_N2: Linear(48, 32), BatchNorm1d(32), LeakyReLu, Linear(32, 8), Dropout(0.1), Sigmoid.
(3) MLP\_N3: Linear(12, 16), BatchNorm1d(16), LeakyReLu.
(4) MLP\_N4: Linear(80, 64), BatchNorm1d(64), LeakyReLu.

\subsection{Evaluation Metrics}

To measure the accuracy of the ALT predictions, we adopted the commonly used metrics RMSE (Root Mean Square Error) \eqref{RMSE}, MAE (Mean Absolute Error) \eqref{MAE}, MAPE (Mean Absolute Percentage Error) \eqref{MAPE}, as well as the defined $BadRatio_{\gamma}$ \eqref{BadRatio}, which are described below.

\begin{equation}\label{RMSE}
  RMSE = \sqrt{\frac{1}{n}\sum_{i=1}^{n}|y_i-\hat{y}_i|^2}
\end{equation}
\begin{equation}\label{MAE}
  MAE = \frac{1}{n}\sum_{i=1}^{n}|y_i-\hat{y}_i|
\end{equation}

\begin{equation}\label{MAPE}
 MAPE = \frac{1}{n}\sum_{i=1}^{n}\frac{|y_i-\hat{y}_i|}{y_i}
\end{equation}

\begin{equation}\label{BadRatio}
 BadRatio_{\gamma} = \frac{\sum_{i=1}^{n}\mathcal{\sigma}(\frac{|y_i-\hat{y}_i|}{y_i}>\gamma)}{n}
\end{equation}

Here, $n$ is the number of test samples, $y_i$ is the ground truth label for the test sample $i$, and $\hat{y}_i$ is the corresponding estimation. For all four metrics, lower values denote better estimations. RMSE is sensitive to the outliers and MAE is actually the L1 loss that directly represents the average estimation errors. MAPE \eqref{MAPE} helps generally evaluate the estimations in terms of large ranges of ground truth values. $BadRatio_{\gamma}$ is to evaluate the robustness of the model, where $\sigma(x)$ is a sign indicator function, i.e.,  $\sigma(x)=1$ if the expression $x$ is true, otherwise 0. $\gamma$ is a manually set threshold value, e.g., $\gamma=0.3$ represents that the absolute percentage errors $\frac{|y_i-\hat{y}_i|}{y_i}$ are higher than $30 \%$, i.e., the relative accuracy would be lower than  $70 \%$. Hereafter, the metric $BadRatio_{\gamma}$ denotes the ratio of bad predictions over the total number of test samples, Hence, a lower $BadRatio_{\gamma}$ presents a better robustness of the model.

\subsection{Result Analysis}
In the experiment study, we remove the "Holding Featurization" as a baseline to testify its effect on the ALT prediction. Additionally, the predictions are evaluated on different settings of the window size parameters $\tau = 60, 90$ (in seconds) and $\delta = 10, 15, 20, 25, 30$ (in minutes). Since the results of $\tau=30$ are quite worse, the corresponding results are not shown. Measured by the metrics of RMSE, MAE, MAPE and $BadRatio_{\gamma}$, the experimental results are shown in \ref{etapred}, where $\gamma=0.3$. The predictions measured by all metrics in \ref{etapred} show that both the baseline and the proposed method performs better with $\tau=60$. Meanwhile, the proposed method performs better than the baseline on all metrics ($\tau=60, \delta=10$).

\begin{figure}[t]
   \begin{center}
	\includegraphics[width=0.75\columnwidth]{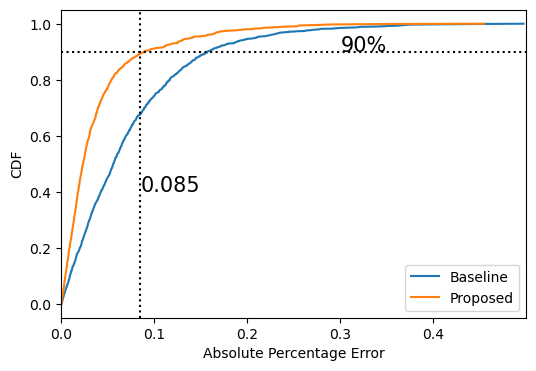}
	\caption{CDF of Absolute Percentage Error. $\tau=60$, $\delta=10$.}
	\label{etaErr_distr}
 \end{center}
\end{figure} 

To further compare the performance of the baseline and the proposed method, we calculate the Absolute Percentage Error (APE) as in Eq. \ref{absErr_ratio}. 

\begin{equation}
    APE = \frac{|y_i-\hat{y}_i|}{y_i}
    \label{absErr_ratio}
\end{equation}

\begin{table}[t]
  \begin{center}
    \caption{Performance Improvement}
    \label{cmp}
    \begin{tabular}{@{}lcr@{}} 
    \hline
      $\tau=60$     & ratio of              & bad ratio \\
      $\delta=10$   & $|err|<60$ seconds    & $\gamma=0.3$\\
      \hline
      Baseline      & 51.07\%               & 1.56\%\\
      Proposed      & 79.40\%               & 0.25\%\\
      Improvement   & 28.37\%               & 1.31\%\\
      \hline
    \end{tabular}
  \end{center}
\end{table}

The cumulative distribution function (CDF) of the APE is shown in Fig. \ref{etaErr_distr}. We can find that the proposed ALT prediction method has quite lower APE values on all cumulative percentages, and it converges to 1 much faster than the baseline. For the proposed ALT prediction method, 90\% predictions are of lower than 85\% prediction errors.

In addition, as shown in Table. \ref{cmp}, the $Bad Ratio$ of the proposed method is improved by 1.31\% compared to the baseline with $\gamma=0.3$. For the proposed method, 79.4\% prediction errors are lower than 60 seconds, which has an 28.37\% improvement compared to the baseline, where the prediction error is $|err| = |y_i-\hat{y}_i|$.

\section{Conclusions}
\label{conclusion}
In this paper, we propose a novel approach for predicting aircraft landing time (ALT) based on deep convolution neural networks applied to the generated trajectory images. The trajectory images capture various information about the aircraft, such as its spatial position, speed, heading, entry zone, relative distances to other aircraft, and real-time arrival traffic flows in the research airspace. These images can replace most of the complex feature engineering, and allow us to use state-of-the-art CNNs for the ALT prediction problem. Moreover, since the aircraft holding stage can significantly affect the ALT, our method use a module for the automatic holding-related feature engineering, which takes the generated images, as well as the leading aircraft holding status, the time and speed gap with the leading aircraft at TRC, and the speed variation at TRC as its input. Its output is further fed into the ALT prediction, and finally an end-to-end ALT prediction model is proposed. We conduct our study on the Changi Airport and use the related ADS-B dataset, as well as the METAR, flight plan, and online APD data in our experiment. Results show that our method can achieve an average accuracy of 96.1\%, with 79.4\% of the prediction errors being lower than 60 seconds. Experimental findings also verified that by integrating the holding featurization module with holding-related features, we can significantly improve the final ALT predictions. Compared to the baseline without the holding featurization, the proposed method reduces the mean absolute error (MAE) from 82.23 seconds to 43.96 seconds, and reduces the bad ratio of the predictions from 1.56\% to 0.25\%. 

Predicting ALTs for inbound aircraft is crucial for sequencing them efficiently. In this paper, we only predict the ALT for each aircraft at the research airspace boundary (TRC), which is 50NM to Changi airport in our case study. However, predicting the ALT with longer distances (e.g., 100NM), choice of the Standard Arrival Route (STAR), and actions of the air traffic controllers would be further beneficial for managing arrival aircraft. Therefore, we plan to improve our proposed ALT prediction method by considering these factors in the future.

\section*{Acknowledgment}

This work is supported by the National Research Foundation, Singapore, and the Civil Aviation Authority of Singapore (CAAS), under the Aviation Transformation Programme. (Grant No. ATP\_IOP for
ATM\_I2R\_2). Any opinions, findings and conclusions, or recommendations expressed in this material are those of the authors and do not reflect the views of the National Research Foundation, Singapore, and the Civil Aviation Authority of Singapore. The authors would like to thank all colleagues from CAAS for providing valuable comments and suggestions on this work.

\end{document}